\documentclass[conference]{IEEEtran}
\usepackage{cite}
\usepackage{amsmath,amssymb,amsfonts}
\usepackage{graphicx}
\usepackage{textcomp}
\usepackage{xcolor}

\usepackage{bm}
\usepackage{algorithm}
\usepackage{algorithmicx}
\usepackage{algpseudocode}
\usepackage{stfloats}
\usepackage{cuted}
\usepackage{capt-of}
\usepackage{multirow}
\usepackage{hyperref}

\begin{document}
\pagestyle{plain}
\thispagestyle{plain}

\title{CLAD: Keypoint Matching Using Global Line Correspondences}
\author{\IEEEauthorblockN{Masaki Satoh}
\IEEEauthorblockA{
\textit{Morpho, Inc.}\\
Tokyo, Japan \\
m-satoh@morphoinc.com}
}

\maketitle

\IEEEpeerreviewmaketitle

\suppressfloats[t]

\begin{figure}[t]
    \centering
    \includegraphics[width=0.9\linewidth]{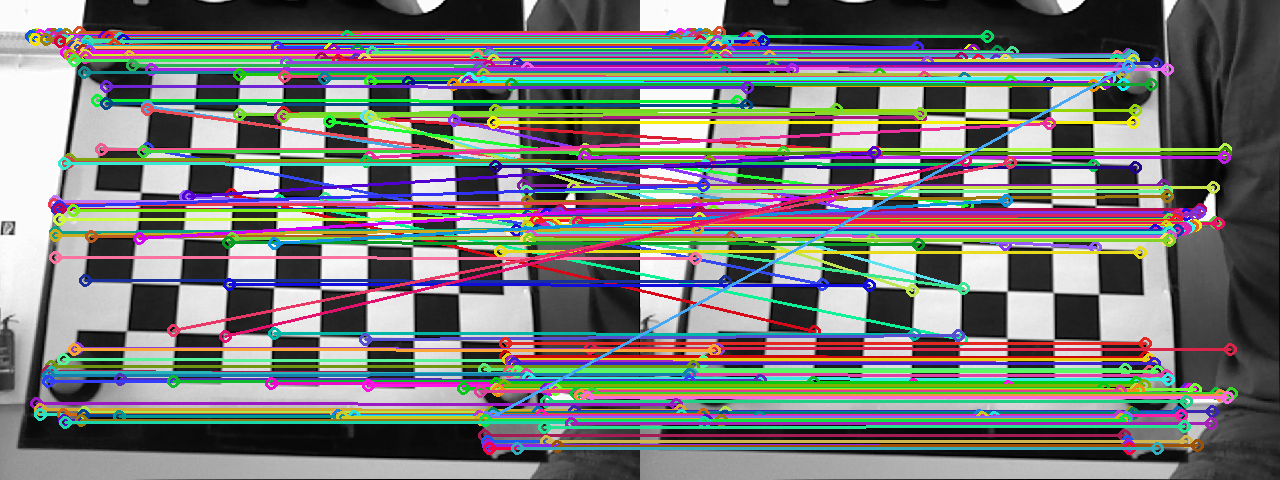}

    \vspace*{0.1cm}

    \includegraphics[width=0.9\linewidth]{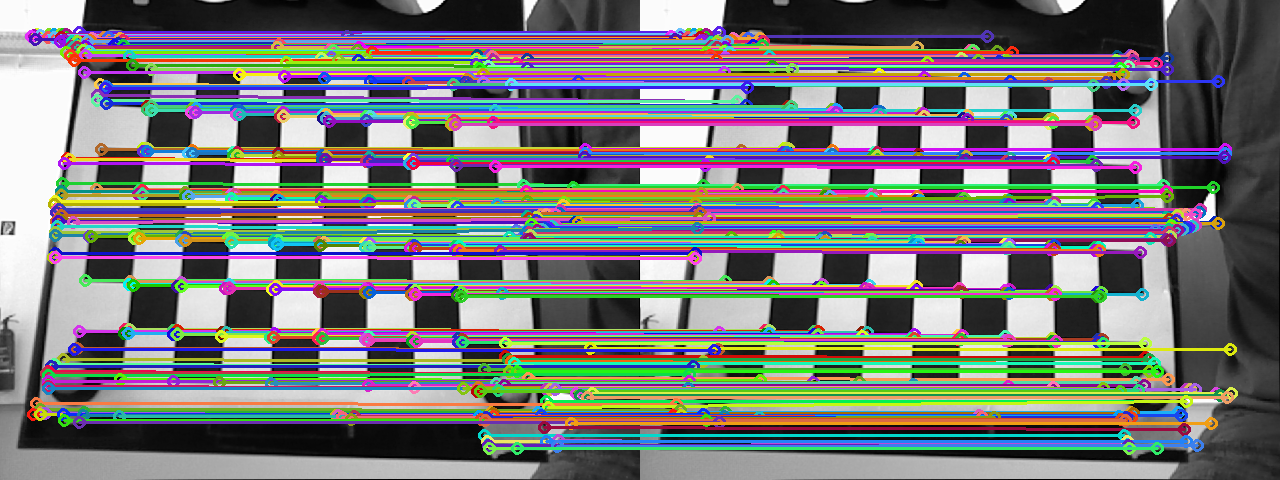}
    \caption{Examples of keypoint matches obtained with ORB (top) and CLAD (bottom). Each pair of corresponding keypoints and its connecting line are shown in the same color. These examples illustrate a limitation of descriptors based only on local image information: in challenging scenes, many local patches may look alike, causing methods such as ORB to confuse different keypoints. In contrast, CLAD uses the global geometric context provided by line correspondences, which helps it find more reliable matches in these cases.}
    \label{fig:keypoint_matches}
\end{figure}

\begin{abstract}
    Traditional keypoint descriptors, such as SIFT (Scale-Invariant Feature Transform) and ORB (Oriented FAST and Rotated BRIEF), are computed from local image patches. They may be unreliable in scenes with repetitive patterns or weak textures, where patches at different locations can look similar. We propose the Contextual Line-Assisted Descriptor (CLAD), a keypoint matching method that addresses this problem by using global geometric context from line correspondences. CLAD represents each keypoint with a vector whose elements indicate its proximity to matched lines. It can therefore use the keypoint's spatial relationships to the lines in the scene. We use HOME (Hough-space One-dimensional Matching of Extrema) to detect both clear and weak linear structures and to establish enough line-to-keypoint associations to represent this context. Evaluations on the TUM RGB-D SLAM dataset and outdoor video sequences show that CLAD provides geometrically consistent matches that complement those obtained with traditional local descriptors. In repetitive scenes, where local descriptors may not distinguish similar patches, CLAD uses global structure to improve matching reliability. The method is computationally lightweight and practical to implement.
\end{abstract}

\section{Introduction}

Keypoint detection and matching are fundamental to many computer vision applications, including image registration, 3D reconstruction, and simultaneous localization and mapping (SLAM). A keypoint detector identifies distinctive, repeatable points in an image, such as corners and blobs. A matching algorithm then finds correspondences between images by comparing the feature vectors, or descriptors, associated with the keypoints. The number of correct matches directly affects the accuracy and robustness of later processing. We can increase this number either by improving the detector to find more repeatable points or by improving the descriptors to make matching more reliable. This paper focuses on the second approach.

In standard methods, such as the Scale-Invariant Feature Transform (SIFT) \cite{lowe2004sift} and Oriented FAST and Rotated BRIEF (ORB) \cite{rublee2011orb}, each keypoint descriptor is computed only from the local image patch around that keypoint. Local information alone may not distinguish keypoints at different locations if their patches look similar, making it difficult to match such keypoints correctly. Even in these cases, human observers can often establish correct correspondences by considering the scene as a whole and the geometric relationships between keypoints. For example, standard descriptors may not distinguish individual checkerboard corners, as illustrated in \figurename~\ref{fig:keypoint_matches}, whereas a human observer can identify them from the overall grid pattern.

We propose CLAD (Contextual Line-Assisted Descriptor), a method that uses line correspondences to capture global geometric structure and improve keypoint matching. The following simplified example illustrates the main idea. Suppose that we establish line correspondences between two images, where line $\bm{\ell}_i$ in the first image corresponds to line $\bm{\ell}_i'$ in the second. Let $N$ be the number of line correspondences. We represent the $j$th keypoint in the first image and the $k$th keypoint in the second image with $N$-dimensional binary vectors $\bm{v}_j$ and $\bm{v}_k'$, respectively. The $i$th element of $\bm{v}_j$ is $1$ if $\bm{\ell}_i$ passes through the $j$th keypoint in the first image and $0$ otherwise. We define $\bm{v}_k'$ in the same way for the $k$th keypoint in the second image. These binary descriptors thus encode the global linear context. We use the Hamming distance to compare CLAD descriptors and match keypoints, as in standard methods. \figurename~\ref{fig:keypoint_matches} shows that CLAD can match keypoints in challenging scenes where local descriptors such as ORB may fail.

\section{Method}

CLAD matches input keypoints by using global context derived from line correspondences. This section first defines the CLAD descriptor and then describes the keypoint detector and the line detection and matching algorithms used in our implementation. In principle, CLAD can be used with any keypoint detector and any line detection and matching algorithm.

\subsection{Contextual Line-Assisted Descriptor}

For $i = 1, 2, \dots, N$, line $\bm{\ell}_i=(a_i, b_i, c_i)^\mathrm{T}$ in the first image corresponds to line $\bm{\ell}'_i=(a'_i, b'_i, c'_i)^\mathrm{T}$ in the second image. The distance from a keypoint $\bm{x}$ in the first image to line $\bm{\ell}_i$ is
\begin{align}
    d_i(\bm{x}) = \frac{|a_i x + b_i y + c_i|}{\sqrt{a_i^2 + b_i^2}}.
\end{align}
For this keypoint, we define an $N$-dimensional binary descriptor vector $\bm{v}(\bm{x})$ as
\begin{align}
    \bm{v}(\bm{x}) \equiv ( \psi(d_1(\bm{x})), \psi(d_2(\bm{x})), \dots, \psi(d_N(\bm{x})) )^\mathrm{T},
\end{align}
where $\psi(d)$ indicates whether the keypoint is near the line: it is $1$ for small distances and $0$ for large distances. We use the parameter $\sigma_\psi$ to set the size of the proximity region, such that $\psi(d)=1$ when $d<\sigma_\psi$ and $\psi(d)=0$ otherwise. For a keypoint $\bm{x}$, we call the set of lines satisfying $\psi(d_i(\bm{x})) = 1$ its \emph{associated lines}. We define the descriptor for a keypoint $\bm{x}'$ in the second image in the same way.

For keypoints $\bm{x}$ and $\bm{x}'$ in the first and second images to match, corresponding lines should have similar proximity relationships to the two keypoints. That is, if line $\bm{\ell}_i$ passes near $\bm{x}$, its counterpart $\bm{\ell}'_i$ should also pass near $\bm{x}'$; if it does not pass near $\bm{x}$, $\bm{\ell}'_i$ should not pass near $\bm{x}'$ either. Thus, a sufficient number of descriptor elements should agree. We compute the Hamming distance between descriptors and use it for keypoint matching. For efficient storage and computation, we pad the descriptors with zeros so that their lengths are multiples of 8.

\subsection{Keypoint Detection}

As noted above, CLAD can be used with any keypoint detector. We consider the Features from Accelerated Segment Test (FAST) \cite{rosten2006machine} for two reasons. First, it is very fast and is therefore suitable for many applications. Second, a single-scale detector such as FAST avoids assigning scale information that CLAD does not need. Multiscale detection, as used by SIFT, is especially useful when images differ substantially in scale. However, CLAD does not require it, and scale estimates may instead introduce noise by placing keypoints close to one another. Although CLAD could also be applied to multiscale keypoints, we leave this for future work. In our implementation, however, we use ORB rather than FAST for keypoint detection. ORB's keypoint detector is based on FAST, so the detected keypoints are similar. Using ORB keypoints also lets us compare ORB and CLAD on the same set of keypoints. We disable ORB's multiscale feature detection.

\subsection{Line Detection and Matching}

For CLAD to match keypoints successfully, we need enough line correspondences between the images. A standard approach to line detection and matching is to detect line segments with the Line Segment Detector (LSD) \cite{von2010lsd} and match them with the Line Band Descriptor (LBD) \cite{zhang2013lbd}. LSD efficiently detects straight line segments in an image, and LBD computes a robust descriptor for each segment and uses it to find matches across images. For CLAD, we can extend LSD-detected segments to infinite lines and use their LBD correspondences. LSD and LBD are suitable for many applications, including SLAM. However, they are not ideal for CLAD because the detected segments represent only a limited part of the scene's global linear context. In addition, LSD can be computationally expensive, which may limit its use in real-time applications.

Recently, Hough-space-based methods have been proposed for line detection and matching \cite{satoh2026hough,satoh2026home}. One such method is HOME \cite{satoh2026home}. It is lightweight compared with LSD and LBD and is therefore suitable for real-time applications. HOME is based on the texture-preserving Hough transform \cite{satoh2026hough}, a continuous version of the Hough transform. Unlike methods that detect only visible line segments, this transform accumulates image information along lines. It can therefore respond to weak linear structures even when they are not visible as continuous line segments. HOME can thus capture weak and implicit structures that may help represent global context in CLAD. We therefore use HOME in this paper. In Section~\ref{sec:experiments}, we compare LSD and HOME in terms of how much line-based context each method provides to CLAD.

\section{Experiments}
\label{sec:experiments}

The main purpose of our experiments is to show that CLAD provides information that complements traditional keypoint descriptors, such as ORB and SIFT, in keypoint matching. We do not claim that CLAD can replace these descriptors.

Before describing the experiments, we specify the parameters used in our method. For ORB, we use OpenCV's default parameters, except that we disable multiscale feature detection. For HOME, we use the parameters from the original paper. We set the proximity-region size to $\sigma_\psi = 4$. To improve matching reliability, we use Lowe's ratio test \cite{lowe2004sift} with a threshold of $0.75$ for ORB, HOME, and CLAD matching. For each keypoint in the first image, we apply the ratio test to its best and second-best matches in the second image.

\begin{figure}[!htp]
    \centering
    \begin{minipage}{0.31\linewidth}
        \centering
        \includegraphics[width=\linewidth]{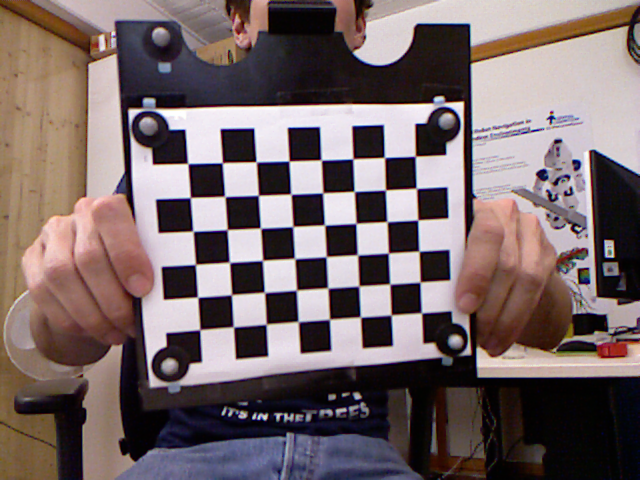}
    \end{minipage}
    \begin{minipage}{0.31\linewidth}
        \centering
        \includegraphics[width=\linewidth]{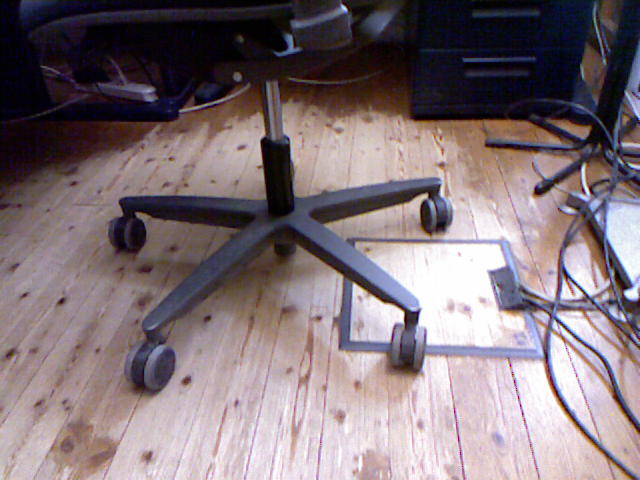}
    \end{minipage}
    \begin{minipage}{0.31\linewidth}
        \centering
        \includegraphics[width=\linewidth]{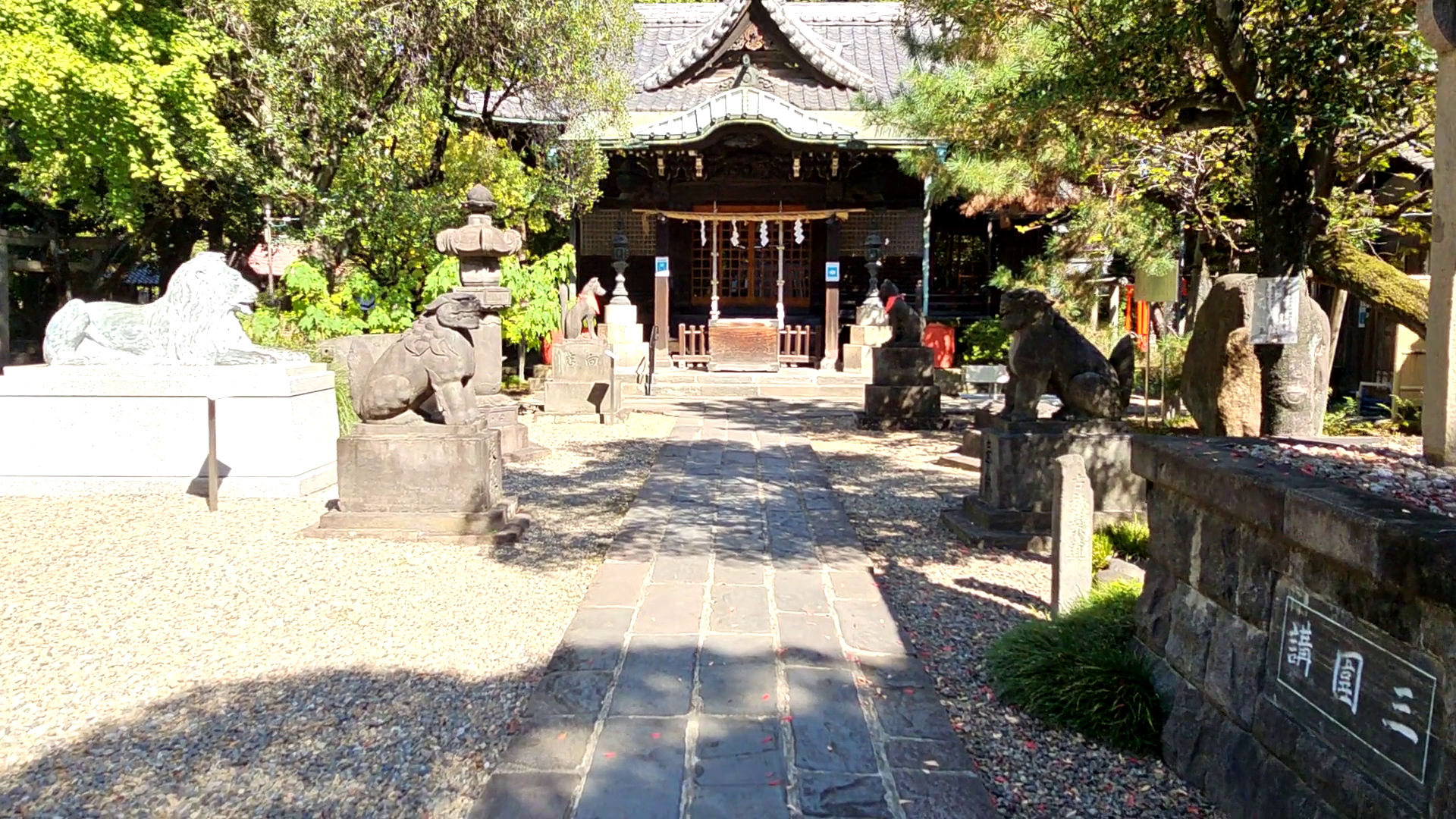}
    \end{minipage}
    \caption{Example frames from the checkerboard (left), indoor (middle), and outdoor (right) sequences.}
    \label{fig:example_images}
\end{figure}

Because HOME is designed to process video frames, we use video sequences in our experiments. First, to illustrate the importance of global context, we use three checkerboard sequences from the TUM RGB-D SLAM dataset \cite{sturm12iros}: ``freiburg1\_rgb\_calibration,'' ``freiburg2\_rgb\_calibration,'' and ``freiburg3\_calibration\_rgb\_depth.'' These sequences are listed in the ``Calibration Files'' category; we refer to them as the checkerboard sequences. Second, to represent a typical use case, we use three indoor sequences from the ``Handheld SLAM'' category of the TUM dataset: ``freiburg1\_360,'' ``freiburg1\_floor,'' and ``freiburg1\_desk.'' We refer to these as the indoor sequences. Third, as a sanity check, we prepare seven texture-rich outdoor sequences captured by smartphones and refer to them as the outdoor sequences. Each outdoor sequence is a four-second video recorded at 30 frames per second (120 frames total) at a resolution of $640 \times 360$ pixels. Sample frames are shown in \figurename~\ref{fig:example_images}.

\subsection{LSD vs HOME}

For CLAD to work effectively, each frame must contain enough detected lines, and each keypoint must have enough associated lines. Because only some detected lines can be matched across images, the number of line correspondences available to CLAD is lower than the total number of detected lines. It is therefore important for each frame to have many detected lines and for each keypoint to have many associated line candidates. We therefore compare LSD, using OpenCV's default settings, with HOME on the checkerboard, indoor, and outdoor sequences. For CLAD, we extend each LSD-detected line segment to an infinite line. We measure the average number of detected lines per frame and the average number of associated line candidates per keypoint. The results are reported in \tablename~\ref{tab:lsd_vs_home}. In the table, ``Keypoints,'' ``Detected,'' and ``Associated'' denote the average numbers of detected ORB keypoints, detected lines, and associated line candidates, respectively. HOME consistently detects more lines and provides more associated line candidates than LSD for all three sequence types. This result supports our choice of HOME as the source of line features for CLAD. HOME reaches its configured maximum of $2000$ lines in most sequences.

\begin{table}[!htb]
    \centering
    \caption{Average numbers of detected lines and associated line candidates for LSD and HOME.}
    \label{tab:lsd_vs_home}
    \begin{tabular}{l|rr|rr|rr}
        & \multicolumn{2}{c|}{Checkerboard} & \multicolumn{2}{c|}{Indoor} & \multicolumn{2}{c}{Outdoor} \\
        & \multicolumn{2}{c|}{($4799$ frames)} & \multicolumn{2}{c|}{($2611$ frames)} &
        \multicolumn{2}{c}{($840$ frames)} \\
        & LSD & HOME & LSD & HOME & LSD & HOME \\
        \hline
        Keypoints &
        \multicolumn{2}{c|}{$497.5$} &
        \multicolumn{2}{c|}{$378.4$} &
        \multicolumn{2}{c}{$500$} \\
        Detected &
        $825.7$ & $1999.9$ &
        $629.7$ & $1985.3$ &
        $929.7$ & $2000$ \\
        Associated &
        $20.1$ & $43.2$ &
        $15.9$ & $41.2$ &
        $19.7$ & $42.8$ \\
    \end{tabular}
\end{table}

\subsection{ORB vs CLAD}

We compare CLAD and ORB by matching pairs of video frames. For the checkerboard and indoor sequences, we match adjacent frames. Because adjacent-frame matching is too easy for the outdoor sequences, we instead match pairs of outdoor frames separated by four frames. Ground-truth correspondences for ORB keypoints are difficult to obtain in real-world scenes. We therefore use a practical evaluation to test whether CLAD provides matches that differ from those found by ORB. Because ORB and CLAD use the same keypoints, we expect some of their matches to overlap. We consider overlapping matches more reliable than non-overlapping matches because the two methods use fundamentally different information. We compute a fundamental matrix from the overlapping matches and use it to evaluate the two-view geometric consistency of the overlapping, ORB-only, and CLAD-only matches. Although the estimated matrix is not guaranteed to be perfect, it provides a reasonable basis for comparing the statistical properties of these three match sets. We also evaluate the overlapping matches to check that they are reliable or at least consistent with the two-view geometry. For this experiment, we measure the distances from keypoints to the corresponding epipolar lines in both images. For inlier thresholds $\theta = 1, 3, 9$, we report the average number and ratio of inlier matches, the root-mean-square error (RMSE) of the residuals, and the average total number of matches in each category. We consider a match inlier if the distances are below the threshold $\theta$ in both images. We skip a frame pair if it has fewer than 16 overlapping matches or if fundamental matrix estimation fails. We estimate the fundamental matrix using OpenCV's \texttt{findFundamentalMat} function with the LMEDS algorithm.

The comparison results are shown in \tablename~\ref{tab:orb_vs_clad}, where ``ORB-o'' and ``CLAD-o'' denote ORB-only and CLAD-only matches, respectively. Overall, the overlapping matches have higher inlier ratios and lower RMSE values, indicating that matches found by both ORB and CLAD are consistent with the two-view geometry. The checkerboard sequences most clearly demonstrate the value of global context. The average total number of ORB-only matches is $52.3$, much lower than the $198.3$ CLAD-only matches. Moreover, even at the widest threshold, $\theta=9$, the inlier ratio of the ORB-only matches is only $54.7\%$. This result shows that ORB has difficulty distinguishing locally similar patterns. In contrast, the CLAD-only matches achieve an inlier ratio of $99.9\%$ at the same threshold. Thus, CLAD can use global context to match keypoints reliably, even when their local patterns are ambiguous. \figurename~\ref{fig:keypoint_matches} shows example ORB and CLAD matches from the checkerboard sequences. The example illustrates how ORB can confuse locally similar patterns, whereas CLAD can distinguish keypoints using the global grid structure.

\begin{table*}[!tb]
    \centering
    \caption{Comparison of keypoint matching performance between ORB and CLAD for different threshold values $\theta$.}
    \label{tab:orb_vs_clad}
    \begin{tabular}{ll|rrr|rrr|rrr}
        && \multicolumn{3}{|c}{Checkerboard} & \multicolumn{3}{|c}{Indoor}
        & \multicolumn{3}{|c}{Outdoor} \\
        && \multicolumn{3}{|c}{($4796$ pairs)} & \multicolumn{3}{|c}{($2472$ pairs)}
        & \multicolumn{3}{|c}{($812$ pairs)} \\
        &&
        Overlapped & ORB-o & CLAD-o &
        Overlapped & ORB-o & CLAD-o &
        Overlapped & ORB-o & CLAD-o \\
        \hline
        \hline
        & Total &
        $233.1$ & $52.3$ & $198.3$ &
        $170.2$ & $35.5$ & $95.4$ &
        $233.2$ & $47.1$ & $109.0$ \\
        \hline
        \multirow{3}{*}{Inlier}
        & $\theta=1$ &
        $196.7$ & $15.6$ & $116.9$ &
        $136.6$ & $20.7$ & $41.9$ &
        $200.3$ & $34.3$ & $52.6$ \\
        & $\theta=3$ &
        $228.8$ & $23.5$ & $183.1$ &
        $168.9$ & $31.2$ & $81.1$ &
        $232.3$ & $45.2$ & $96.0$ \\
        & $\theta=9$ &
        $233.0$ & $28.6$ & $198.2$ &
        $170.1$ & $32.9$ & $94.8$ &
        $233.1$ & $46.1$ & $108.0$ \\
        \hline
        \multirow{3}{*}{Inlier ratio}
        & $\theta=1$ &
        $84.4\%$ & $29.9\%$ & $59.0\%$ &
        $80.3\%$ & $58.4\%$ & $43.9\%$ &
        $85.9\%$ & $72.8\%$ & $48.2\%$ \\
        & $\theta=3$ &
        $98.2\%$ & $45.0\%$ & $92.3\%$ &
        $99.3\%$ & $88.1\%$ & $85.0\%$ &
        $99.6\%$ & $95.9\%$ & $88.1\%$ \\
        & $\theta=9$ &
        $100.0\%$ & $54.7\%$ & $99.9\%$ &
        $100.0\%$ & $92.7\%$ & $99.3\%$ &
        $100.0\%$ & $97.9\%$ & $99.1\%$ \\
        \hline
        \multirow{3}{*}{RMSE}
        & $\theta=1$ &
        $0.280$ & $0.396$ & $0.484$ &
        $0.443$ & $0.514$ & $0.539$ &
        $0.406$ & $0.484$ & $0.531$ \\
        & $\theta=3$ &
        $0.628$ & $1.091$ & $1.063$ &
        $0.761$ & $1.061$ & $1.357$ &
        $0.622$ & $0.868$ & $1.281$ \\
        & $\theta=9$ &
        $0.855$ & $2.488$ & $1.572$ &
        $0.830$ & $1.477$ & $2.076$ &
        $0.675$ & $1.080$ & $1.847$ \\
    \end{tabular}

    \vspace*{1.5\baselineskip}

    \caption{Comparison of keypoint matching performance between ORB+SIFT' and CLAD for different threshold values $\theta$.}
    \label{tab:orb+_vs_clad}
    \begin{tabular}{ll|rrr|rrr|rrr}
        && \multicolumn{3}{|c}{Checkerboard} & \multicolumn{3}{|c}{Indoor}
        & \multicolumn{3}{|c}{Outdoor} \\
        && \multicolumn{3}{|c}{($4796$ pairs)} & \multicolumn{3}{|c}{($2472$ pairs)}
        & \multicolumn{3}{|c}{($812$ pairs)} \\
        &&
        Overlapped & ORB+S'-o & CLAD-o &
        Overlapped & ORB+S'-o & CLAD-o &
        Overlapped & ORB+S'-o & CLAD-o \\
        \hline
        \hline
        & Total &
        $254.8$ & $101.2$ & $176.5$ &
        $195.6$ & $52.4$ & $67.9$ &
        $264.2$ & $65.7$ & $77.9$ \\
        \hline
        \multirow{3}{*}{Inlier}
        & $\theta=1$ &
        $209.6$ & $19.8$ & $106.5$ &
        $151.6$ & $27.0$ & $28.6$ &
        $219.2$ & $41.5$ & $36.6$ \\
        & $\theta=3$ &
        $249.7$ & $34.0$ & $165.0$ &
        $192.8$ & $44.5$ & $55.7$ &
        $262.2$ & $61.4$ & $66.9$ \\
        & $\theta=9$ &
        $254.8$ & $44.8$ & $176.4$ &
        $195.5$ & $48.3$ & $67.3$ &
        $264.2$ & $64.6$ & $77.1$ \\
        \hline
        \multirow{3}{*}{Inlier ratio}
        & $\theta=1$ &
        $82.3\%$ & $19.5\%$ & $60.3\%$ &
        $77.5\%$ & $51.4\%$ & $42.1\%$ &
        $83.0\%$ & $63.1\%$ & $46.9\%$ \\
        & $\theta=3$ &
        $98.0\%$ & $33.6\%$ & $93.5\%$ &
        $98.6\%$ & $84.9\%$ & $82.0\%$ &
        $99.2\%$ & $93.4\%$ & $85.8\%$ \\
        & $\theta=9$ &
        $100.0\%$ & $44.2\%$ & $99.9\%$ &
        $100.0\%$ & $92.2\%$ & $99.1\%$ &
        $100.0\%$ & $98.3\%$ & $98.9\%$ \\
        \hline
        \multirow{3}{*}{RMSE}
        & $\theta=1$ &
        $0.289$ & $0.433$ & $0.480$ &
        $0.453$ & $0.522$ & $0.537$ &
        $0.415$ & $0.499$ & $0.530$ \\
        & $\theta=3$ &
        $0.673$ & $1.256$ & $1.023$ &
        $0.835$ & $1.193$ & $1.376$ &
        $0.709$ & $1.068$ & $1.290$ \\
        & $\theta=9$ &
        $0.892$ & $2.792$ & $1.472$ &
        $0.948$ & $1.717$ & $2.220$ &
        $0.779$ & $1.384$ & $1.954$ \\
    \end{tabular}
\end{table*}

The ORB-only and CLAD-only matches also show different characteristics on the indoor and outdoor sequences. CLAD-only matches are more numerous than ORB-only matches, but the ORB-only matches have lower residuals. The ORB-only inlier ratio is higher than the CLAD-only inlier ratio at strict thresholds, such as $\theta=1$, while the ordering reverses at the relaxed threshold of $\theta=9$. In other words, CLAD finds more matches, but some are close to, rather than exactly on, the correct correspondences. This behavior is consistent with the expected characteristics of global and local methods.

\subsection{ORB+SIFT' vs CLAD}

Our goal is to show that CLAD provides matches that complement traditional methods based on local descriptors. A comparison with other methods, such as SIFT, would therefore be useful. However, this comparison is difficult because each method normally uses its own keypoint detector. A meaningful comparison requires fixing the keypoint set and varying only the matching algorithm. To make this comparison, we compute SIFT descriptors at the keypoints detected by ORB and match them using Lowe's ratio test with a threshold of $0.75$, as for ORB and CLAD. We use OpenCV's SIFT implementation with its default parameters and set the keypoint size to 5 pixels. We then combine these pseudo-SIFT matches with the standard ORB matches and compare the result with CLAD. We call this combination ORB+SIFT'. The results are shown in \tablename~\ref{tab:orb+_vs_clad}, where ``ORB+S'-o'' denotes matches found only by ORB+SIFT'. Adding pseudo-SIFT matches increases the numbers of overlapping and ORB+SIFT'-only matches and decreases the number of CLAD-only matches. However, the overall trend is similar to that in the ORB-versus-CLAD comparison.

These results show that CLAD provides matches that complement traditional local-feature matching methods. Although ORB generally provides more accurate matches, CLAD can find additional, globally consistent matches that ORB alone may miss.

\subsection{Processing Time}

Finally, we report the processing time of the current CLAD implementation, measured on a laptop with an AMD Ryzen 7 7735U CPU and 32 GB of RAM. We use the outdoor sequences because they contain relatively many ORB and HOME features and thus provide a demanding test of processing time. The average processing times are 1.580 ms for ORB keypoint detection, 3.919 ms for HOME keypoint detection and descriptor extraction, 4.321 ms for HOME matching, 3.545 ms for CLAD descriptor extraction, and 1.072 ms for CLAD matching. Although we have not thoroughly optimized the implementation, these results indicate that it can operate at practical processing rates and may run faster with further optimization.

\section{Conclusion}

In this paper, we introduced CLAD, a keypoint matching method that uses line correspondences to incorporate global context and complement traditional local-feature matching methods. Line correspondences capture linear structures in a scene and can provide geometric information, such as a homography. However, they have fewer applications than point correspondences; for example, line correspondences alone cannot be used to compute the fundamental matrix \cite{hartley2003multiple}. CLAD associates this global context with local keypoints detected by traditional methods, making global linear context available to standard keypoint-matching pipelines. This design may allow CLAD to be used in a wide range of applications.

Several directions remain for future research. The most important next step is to evaluate CLAD in computer vision tasks such as 3D reconstruction and visual localization. Another direction is to improve HOME or evaluate other line-based methods to further enhance CLAD. Because HOME is designed to process video frames, improving or replacing it could extend CLAD to general keypoint-matching tasks. FAST and other traditional detectors are designed for local-feature matching and may not be optimal for CLAD. Developing or adapting keypoint detectors better suited to CLAD could therefore further improve its performance. Because CLAD provides information that complements traditional local-feature matching methods, designing descriptors that combine local and global information could further improve matching performance.

\section*{Acknowledgment}

The author thanks Takeshi Miura for fruitful discussions and Masaki Hilaga for his support and for permitting this research as part of the author's corporate duties. The author used OpenAI's ChatGPT, Google Gemini, and GitHub Copilot to edit the language, organize the manuscript, and improve its presentation. The author reviewed and verified all AI-assisted content and is solely responsible for the final manuscript.

\bibliographystyle{IEEEtran}
\bibliography{references}

\end{document}